\documentclass[11pt,a4paper]{article}
\usepackage[hyperref]{acl2019}
\usepackage{times}
\usepackage{latexsym}
\usepackage{lipsum}
\usepackage{amsmath}
\usepackage{amsfonts}
\usepackage{algorithm}
\usepackage{graphicx}
\usepackage[noend]{algpseudocode}
\usepackage{color, colortbl}

\usepackage{url}

\definecolor{color1}{HTML}{da6752}
\definecolor{color2}{HTML}{5573a6}
\definecolor{color3}{HTML}{6f9f6a}
\definecolor{color4}{HTML}{f3905c}

\definecolor{Gray}{gray}{0.9}

\newcommand{\shortarrow}[1][5pt]{\mathrel{%
  \vcenter{\hbox{\rule[-.2pt]{#1}{.4pt}}}%
  \mkern-4mu\hbox{\usefont{U}{lasy}{m}{n}\symbol{41}}}}
\newcommand{\veryshortarrow}{\ {\shortarrow}\:}

\aclfinalcopy 

\title{Simple and Effective Curriculum Pointer-Generator Networks for Reading Comprehension over Long Narratives}

\author{$^1$Yi Tay, $^2$Shuohang Wang, $^3$Luu Anh Tuan, $^4$Jie Fu, $^5$Minh C. Phan\\ \textbf{$^6$Xingdi Yuan, $^7$Jinfeng Rao, $^8$Siu Cheung Hui, $^{9}$Aston Zhang}\\
  $^{1,5,8}$Nanyang Technological University \:\:\:\: $^{2}$Singapore Management University \:\:\:\:$^{3}$MIT CSAIL \\  $^{4}$Mila, Polytechnique Montr\'eal \:\:\:\:
  $^{6}$Microsoft Research, Montr\'eal 
  \:\: $^{7}$Facebook \:\: $^{8}$Amazon AI \\
  }

\date{}

\begin{document}
\maketitle
\begin{abstract}
This paper tackles the problem of reading comprehension over long narratives where documents easily span over thousands of tokens. We propose a curriculum learning (CL) based Pointer-Generator framework for reading/sampling over large documents, enabling diverse training of the neural model based on the notion of alternating contextual difficulty. This can be interpreted as a form of domain randomization and/or generative pretraining during training. To this end, the usage of the Pointer-Generator softens the requirement of having the answer within the context, enabling us to construct diverse training samples for learning. Additionally, we propose a new Introspective Alignment Layer (IAL), which reasons over decomposed alignments using block-based self-attention. We evaluate our proposed method on the NarrativeQA reading comprehension benchmark, achieving state-of-the-art performance, improving existing baselines by $51\%$ relative improvement on BLEU-4 and $17\%$ relative improvement on Rouge-L. Extensive ablations confirm the effectiveness of our proposed IAL and CL components.
\end{abstract}

\section{Introduction}
Teaching machines to read and comprehend is a fundamentally interesting and challenging problem in AI research \cite{hermann2015teaching,trischler2016newsqa,rajpurkar2016squad}. While there have been considerable and broad improvements in reading and understanding textual snippets, the ability for machines to read/understand complete stories and novels is still in infancy \cite{kovcisky2018narrativeqa}. The challenge becomes insurmountable in lieu of not only the large context but also the intrinsic challenges of narrative text which arguably requires a larger extent of reasoning. As such, this motivates the inception of relevant, interesting benchmarks such as the NarrativeQA Reading Comprehension challenge\footnote{We tackle the full story setting instead of the summary setting which, inherently, is a much harder task.} \cite{kovcisky2018narrativeqa}. 

The challenges of having a long context have been traditionally mitigated by a two-step approach - retrieval first and then reading second \cite{chen2017reading,wang2018r,lin2018denoising}. This difficulty mirrors the same challenges of open domain question answering, albeit introducing additional difficulties due to the nature of narrative text (stories and retrieved excerpts need to be coherent). While some recent works have proposed going around by training retrieval and reading components end-to-end, this paper follows the traditional paradigm with a slight twist. We train our models to be robust regardless of whatever is retrieved. This is in similar spirit to domain randomization \cite{tobin2017domain}.

In order to do so, we propose a diverse curriculum learning scheme \cite{bengio2009curriculum} based on two concepts of difficulty.  The first, depends on whether the answer exists in the context (\textit{answerability}), aims to bridge the gap between training time and inference time retrieval. On the other hand, and the second, depends on the size of retrieved documents (coherence and \textit{understandability}). While conceptually simple, we found that these heuristics help improve performance of the QA model. To the best of our knowledge, we are the first to incorporate these notions of difficulty in QA reading models.

All in all, our model tries to learn to generate the answer even if the correct answer does not appear as evidence which acts as a form of \textit{generative pretraining during training}. As such, this is akin to learning to guess, largely motivated by how humans are able to extrapolate/guess even when given access to a small fragment of a film/story. In this case, we train our model to generate answers, making do with whatever context it was given. To this end, a curriculum learning scheme controls the extent of difficulty of the context given to the model.

At this juncture, it would be easy to realize that standard pointer-based reading comprehension models would not adapt well to this scheme, as they fundamentally require the golden label to exist within the context \cite{wang2016machine, seo2016bidirectional}. As such, our overall framework adopts a pointer-generator framework \cite{see2017get} that learns to point and generate, conditioned on not only the context but also the question. This relaxes this condition, enabling us to train our models with diverse views of the same story which is inspired by domain randomization \cite{tobin2017domain}. For our particular task at hand, the key idea is that, even if the answer is not found in the context, we learn to generate the answer despite the noisy context. 

Finally, our method also incorporates a novel Introspective Alignment Layer (IAL). The key idea of the IAL mechanism is to introspect over decomposed alignments using block-style local self-attention. This not only imbues our model with additional reasoning capabilities but enables a finer-grained (and local-globally aware) comparison between soft-aligned representations. All in all, our IAL mechanism can be interpreted as learning a matching over matches.

\paragraph{Our Contributions}
All in all, the prime contributions of this work is summarized as follows:
\begin{itemize}
    \item We propose a curriculum learning based Pointer-Generator model for reading comprehension over narratives (long stories). For the first time, we propose two different notions of difficulty for constructing diverse views of long stories for training. We show that this approach achieves better results than existing models adapted for open-domain question answering.
    \item Our proposed model incorporates an Introspective Alignment Layer (IAL) which uses block-based self-attentive reasoning over decomposed alignments. Ablative experiments show improvements of our IAL layer over the standard usage of vanilla self-attention. 
    \item Our proposed framework (IAL-CPG) achieves state-of-the-art performance on the NarrativeQA reading comprehension challenge. On metrics such as BLEU-4 and Rouge-L, we achieve a $17\%$ relative improvement over prior state-of-the-art and a \textbf{10} times improvement in terms of BLEU-4 score over BiDAF, a strong span prediction based model. 
    \item We share two additional contributions. Firstly, we share negative results on using Reinforcement Learning to improve the quality of generated answers \cite{paulus2017deep,bahdanau2016actor}. Secondly, we show that the evaluation scheme in NarrativeQA is flawed and models can occasionally generate satisfactory (correct) answers but score zero points during evaluation. 
\end{itemize}

\section{Our Proposed Framework}
\begin{figure}
    \centering
  \includegraphics[width=0.9\linewidth]{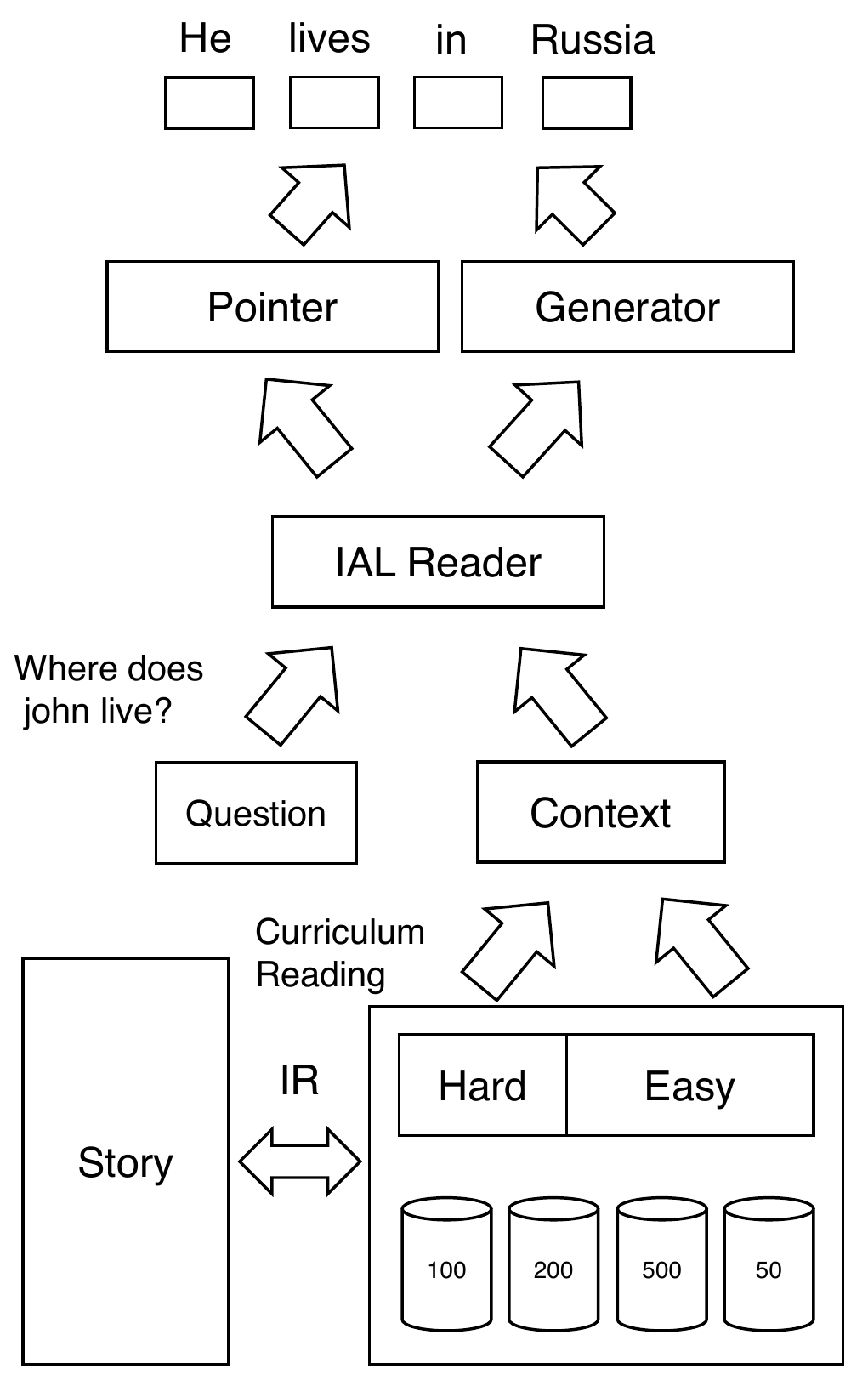}
    \caption{Illustration of our proposed IAL-CPG framework.}
    \label{fig:illustration}
\end{figure}

This section outlines the components of our proposed architecture. Since our problem is mainly dealing with extremely long sequences, we employ an initial retrieval\footnote{This is unavoidable since supporting up to 20K-30K words in computational graphs is still not manageable even with top-grade GPUs.} phrase by either using the answer or question as a cue (query for retrieving relevant chunks/excerpts). The retrieval stage is controlled by our curriculum learning process in which the details are deferred to subsequent sections. The overall illustration of this framework is depicted in Figure \ref{fig:illustration}.
\subsection{Introspective Alignment Reader} 
This section introduces our proposed Introspective Alignment Reader (IAL-Reader). 
\paragraph{Input and Context Encoding} Our model accepts two inputs, (context $C$ and question $Q$). Each input is a sequence of words. We pass each sequence into a shared Bidirectional LSTM layer. 
\begin{align*}
H^c = \text{BiLSTM}(C) \:\:\:\text{,}\:\:\: H^q = \text{BiLSTM}(Q)
\end{align*}
where $H^c \in \mathbb{R}^{\ell_c \times d}$ and $H^q \in \mathbb{R}^{\ell_q \times d}$ are the hidden representations for $C$ and $Q$ respectively.

\paragraph{Introspective Alignment}
Next, we pass $H^c,H^q$ into an alignment layer. Firstly, we compute a soft attention affinity matrix between $H^c$ and $H^q$ as follows:
\begin{equation}
    E_{ij} = F(h_{i}^{c})^{\top}\:F(h_{j}^{q}) \label{align1}
\end{equation}
where $h_{i}^{c}$ is the $i$-th word in the context and $h_{j}^{q}$ is the $j$-th word in the question. $F(\cdot)$ is a standard nonlinear transformation function (i.e., $F(x) = \sigma(Wx + b)$, where $\sigma$ indicates non-linearity function), and is shared between context and question. $E \in \mathbb{R}^{ \ell_c \times \ell_q }$ is the soft matching matrix. To learn alignments between context and question, we compute:
\begin{align*}
A = \text{Softmax}(E)\:H^q    
\end{align*}
where $A \in \mathbb{R}^{\ell_c \times d}$ is the aligned representation of $H^c$. 

\paragraph{Reasoning over Alignments}
Next, to reason over alignments, we compute a self-attentive reasoning over decomposed alignments:
\begin{equation}
\begin{aligned}
\label{sa}
G_{ij} = F_{s}&([A_{i}; H^{c}_{i}; A_{i}-H^c_i, A_{i} \odot H^{c}_{i}])^{\top} \cdot \\ F_{s}&([A_{j}; H^{c}_{j}; A_{j}-H^c_j, A_{j} \odot H^{c}_{j}])
\end{aligned}
\end{equation}
where square brackets $\lbrack \cdot;\cdot \rbrack$ denote vector concatenation, $F_s(\cdot)$ is another nonlinear transformation layer which projects onto $4d$ dimensions. $i$ is the positional index of each word token. Intuitively, $A_i$ comprises of softly aligned question representations with respect to the context. The usage of the \textit{Hadamard} and \textit{Subtraction} operators helps to enhance the degree of comparison/matching. Hence, by including an additional local reasoning over these enhanced alignment vectors, our model can be interpreted as introspecting over alignment matches.

\paragraph{Local Block-based Self-Attention}
Since $\ell_c$ is large in our case (easily $\geq$ 2000), computing the above Equation (\ref{sa}) may become computationally prohibitive. As such, we compute the scoring function for all cases where $|i-j|\leq b$, in which, $b$ is a predefined hyperparameter and also the block size. Intuitively, the initial alignment layer (i.e., Equation \ref{align1}) already considers a global view. As such, this self-attention layer can be considered as a local-view perspective, confining the affinity matrix computation to a local window of $b$. Finally, to compute the introspective alignment representation, we compute:
\begin{align*}
B = \text{Softmax}(G) \: [A; H^c;A - H^c; A\odot H^c]      
\end{align*}
where $B^{\ell_c \times 4d}$ is the introspective aligned representation of $A$. Finally, we use another $d$ dimensional BiLSTM layer to aggregate the aligned representations:
\begin{equation}
Y = \text{BiLSTM}([B;A; H^c;A - H^c; A\odot H^c])    
\end{equation}
where $Y \in \mathbb{R}^{\ell_c \times 2d}$ is the final contextual representation of context $C$.

\subsection{Pointer-Generator Decoder}
Motivated by recent, seminal work in neural summarization, our model adopts a pointer-generator architecture \cite{see2017get}. Given $Y$ (the question infused contextual representation), we learn to either generate a word from vocabulary, or point to a word from the context. The decision to generate or point is controlled by an additive blend of several components such as the previous decoder state and/or question representation. 

The pointer-generator decoder in our framework uses an LSTM decoder\footnote{To initialize the LSTM, we use an additional projection layer over the mean pooled representation of $Y$ similar to \cite{xu2015show}.} with a cell state $c_{t} \in \mathbb{R}^{n}$ and hidden state vector $h_{t} \in \mathbb{R}^{n}$. At each decoding time step $t$, we compute an attention over $Y$ as follows:
\begin{align}
    g_{i} &= \text{tanh}(F_{a}(y_{i}) + F_{h}(h_{t-1}) + F_{q}(H^q)), \\ 
    a_{i} &= g_{i}^{\top}w_{a} \:\:,\:\: y_{t} = \sum^{\ell_c}_{i=0} a_{i} \cdot y_{i}   
\label{eqn:ext_distribution}
\end{align}
where $F_{a}(\cdot)$ and $F_h(\cdot)$ are nonlinear transformations projecting to $n$ dimensions. $i$ is the position index of the input sequence. $F_{q}(\cdot)$ is an additional attentive pooling operator over the question representation $H_q$ (after the context encoding layer). The semantics of the question may be lost after the alignment based encoding. As such, this enables us to revisit the question representation to control the decoder. $y_{t} \in \mathbb{R}^{n}$ is the context representation at decoding time step $t$ and $a \in \mathbb{R}^{\ell_{c}}$ is an attention distribution over the context words which is analogous to the final probability distributions that exist in typical span prediction models. Next, we compute the next hidden state via:
\begin{align*}
h_{t}, c_{t} = \text{LSTM}([y_{t};w_{t-1}], h_{t-1}, c_{t-1})
\end{align*}
where $w_{t-1}$ is the $(t-1)_{th}$ token in the ground truth answer (teacher forcing). To learn to generate, we compute:
\begin{equation}
v_{t} = W_{v}(h_{t}) + b_{v}    \label{eqn:abs_distribution}
\end{equation}
where $v_{t} \in \mathbb{R}^{|V_g|}$, $V_g$ is the global vocabulary size. 
The goal of the pointer-generator decoder is to choose between the abstractive distribution $v_t$ over the vocabulary (see Equation~\ref{eqn:abs_distribution}) and the extractive distribution $a_t$ (see Equation~\ref{eqn:ext_distribution}) over the context text tokens.
To this end, we learn a scalar switch $p_t \in \mathbb{R}$:
\begin{align*}
p_{t} = \text{sigmoid}(F_{pc}(c_{t}) + F_{ph}(h_{t}) + F_{py}(y_{t}))  
\end{align*}
where $F_{pc}(\cdot), F_{ph}(\cdot), F_{py}(\cdot)$ are linear transformation layers (without bias) which project $c_{t},h_{t}$ and $y_{t}$ into scalar values. To control the blend between the attention context and the generated words, we use a linear interpolation between $a_t$ and $v_{t}$. The predicted word $w_{t}$ at time step $t$ is therefore:
\begin{align*}
w_{t} = \text{argmax}(p_{t}\cdot a_t + (1-p_{t}) v_{t})    
\end{align*}
Note that we scale (append and prepend) $a_{t}$ and $v_{t}$ with zeros to make them the same length (i.e., $\ell_{c} + |V_g|$). The LSTM decoder runs for a predefined fix answer length. During inference, we simply use greedy decoding to generate the output answer.

\subsection{Curriculum Reading}
A key advantage of the pointer-generator is that it allows us to generate answers even if the answers do not exist in the context. This also enables us to explore multiple (diverse) views of contexts to train our model. However, to this end, we must be able to identify effectively the most useful retrieved context evidences for the training. For that purpose, we propose to use a diverse curriculum learning scheme which is based on two intuitive notions of difficulty:

\paragraph{Answerability} - It is regarded as common practice to retrieve excerpts based by using the correct answer as a cue (during training). This establishes an additional gap between training and inference since during inference, correct answers are not available. This measure aims to bridge the gap between question and answer (as a query prompt for passage retrieval). In this case, we consider the set of documents retrieved based on questions as the \textit{hard} setting, $H$. Conversely, the set of retrieved documents using answers is regarded as the \textit{easy} setting, $E$. 
\paragraph{Understandability} - This aspect controls how understandable the overall retrieved documents are as a whole. The key idea of this setting is to control the  paragraph/chunk size. Intuitively, a small paragraph/chunk size would enable more relevant components to be retrieved from the document. However, its understandability might be affected if paragraph/chunk size is too small. Conversely, a larger chunk size would be easier to be understood. To control the level of understandability, we pre-define several options of chunk sizes (e.g., $\{50,100,200,500\}$) which will be swapped and determined during training. 

To combine the two measures described above, we comprise an easy-hard set pair for each chunk size, i.e., 
$\{E_k, H_k\}$, where:
\begin{equation}
\begin{aligned}
k &\in \{50, 100, 200, 500\}, \\
E_n &\leftarrow F(\text{corpus}, \text{answer},n), \\
H_n &\leftarrow F(\text{corpus}, \text{question},n)
\end{aligned}
\end{equation}
$F(.)$ is an arbitrary ranking function which may or may not be parameterized, and $n$ is the size of each retrieved chunk. 

\begin{algorithm}[t]
\small
\caption{Curriculum Reading}\label{Alg:CR}
\begin{algorithmic}[1]
\State $chunk\_list \gets \{50,100,200,500\}$
\State $n \gets$ $sample$ $i$ $in$ $chunk\_list$
\State $chunk\_list \gets chunk\_list \setminus \{n\}$
\State $E_n \gets F(Corpus, Answers, n)$
\State $H_n \gets F(Corpus, Questions, n)$
\State $D \gets E_n$ \Comment{initial training set}
\State $count \gets 0$ \Comment{number of swaps within a chunk size}
\For{$i \gets 1$ to
$numEpochs$}       
        \State $Train(D)$
        \State $score \gets Evaluate(Dev\_set)$
        \If{$score<bestDev$}
            \If{$count <= 1/\delta$}
                \State $D \gets Swap(D, E_n, H_n, \delta)$ \Comment{Swap $\delta$ percent of easy set in $D$ with the hard set}
                \State $count \gets count + 1$
            \Else
                \State $Repeat$ $step$ $3$ $to$ $8$ \Comment{Replace training set with new easy set of another chunk size}
            
            \EndIf
        \Else\State $bestDev = score$
        \EndIf
    \EndFor
\end{algorithmic}
\end{algorithm}

 \paragraph{Two-layer Curriculum Reading Algorithm. } As our model utilizes two above measures of difficulty, there lies a question on which whether we should swap one measure at a time or swap both whenever the model meets the failure criterion. In our case, we find that prioritizing answerability over understandability is a better choice. More concretely, at the beginning of the training, we start with an easy set $E_k$ of a random chunk size $k$. When the failure criterion is met (e.g. the model score does not improve on the validation set), we randomly swap a small percent $\delta$ (e.g., 5\% in our experiments\footnote{In early experiments, we found that $5\%-10\%$ works best.}) of the easy set $E_k$ with the hard set $H_k$ within its own chunk size group $k$ to improve the $answerability$. In this case, after $\frac{1}{\delta}$ failures, the model runs out of easy set $E_k$ and is completely based on the hard set $H_k$. At this junction, we swap the model for \textit{understandability}, replacing the training set with a completely new easy set $E_l$ of another chunk size $l$, and repeat the above process. The formal description of our proposed curriculum reading is introduced in Algorithm \ref{Alg:CR}.

\begin{table*}[h]
 \centering

\begin{tabular}{c|c|cccc|cccc}
\hline
 &  & \multicolumn{4}{c}{Dev Set} &  \multicolumn{4}{c}{Test Set} \\
Model & $\ell$ & BLEU-1 & BLEU-4 & Meteor & Rouge & BLEU-1 & BLEU-4 & Meteor & Rouge \\
\hline
IR (BLEU) &  - & 6.73 & 0.30  & 3.58  & 6.73  & 6.52  & 0.34  & 3.35 & 6.45 \\
IR (ROUGE) &  - & 5.78& 0.25 & 3.71 & 6.36 & 5.69 & 0.32  & 3.64 & 6.26 \\
IR (Cosine) &  - & 6.40 & 0.28  & 3.54 & 6.50  & 6.33 & 0.29  &3.28 & 6.43 \\
\hline

BiDAF & - & 5.82 & 0.22 & 3.84 & 6.33 & 5.68 & 0.25 & 3.72 & 6.22 \\
ASR  & 200 & 16.95 & 1.26 & 3.84 & 1.12 & 16.08 & 1.08 & 3.56 & 11.94 \\
ASR & 400 & 18.54 & 0.00 & 4.2 & 13.5 & 17.76 & 1.10 & 4.01 & 12.83 \\
ASR & 1K & 18.91 & 1.37 & 4.48 & 14.47 & 18.36 & 1.64 & 4.24 & 13.4 \\
ASR & 2K & 20.00 & 2.23 & 4.45 & 14.47 & 19.09 & 1.81 & 4.29 & 14.03 \\
ASR  & 4K &19.79 & 1.79 & 4.60 & 14.86 & 19.06 & 2.11 & 4.37 & 14.02 \\
\hline

ASR (Ours) &  4K & 12.03  & 1.06 & 3.10  & 8.87  & 11.26 & 0.65 & 2.66 & 8.68 \\
$R^3$  & - & 16.40 & 0.50 & 3.52 & 11.40 & 15.70 & 0.49 & 3.47 & 11.90 \\
RNET-PG &  4K & 17.74 & 0.00  & 3.95  & 14.56  & 16.89  & 0.00  & 3.84 & 14.35\\
RNET-CPG & 4K & 19.71 & 2.05  & 4.91  & 15.05  & 19.27  & 1.45  & 4.87 & 15.50 \\
\hline 
IAL-CPG  & 4K & \textbf{23.31} &	\textbf{2.70}	&\textbf{5.68} &	 \textbf{17.33}	& \textbf{22.92} & 	\textbf{2.47} &	\textbf{5.59}	&\textbf{17.67}\\
\hline
Rel. Gain & - &  +31\% & +51\% & +23\% & +17\% & +20\% & +17\% & +28\% & +26\% \\
\hline
\end{tabular}
\caption{Results on NarrativeQA reading comprehension dataset (Full story setting). Results are reported from \cite{kovcisky2018narrativeqa} .The numbers besides the model name denote the total context size. Rel. Gain reports the relative improvement of our model and the best baseline reported in \cite{kovcisky2018narrativeqa} on a specific context size setting.}
\label{mainresults}
\end{table*}

\section{Experiments}
We conduct our experiments on the NarrativeQA reading comprehension challenge.
\subsection{Experimental Setup}
This section introduces our experimental setups.
\paragraph{Model Hyperparameters} We implement our model in Tensorflow. Our model is trained with Adadelta \cite{zeiler2012adadelta}. The initial learning rate is tuned amongst $\{0.1, 0.2, 0.5\}$. The L2 regularization is tuned amongst $\{10^{-8},10^{-6},10^{-5}\}$.  The size of the LSTM at the encoder layer is set to $128$ and the decoder size is set to $256$. The block size $b$ for the Introspective Alignment Layer is set to 200. We initialize our word embeddings with pre-trained GloVe vectors \cite{pennington2014glove} which are not updated\footnote{In our early experiments, we also masked entities following the original work \cite{kovcisky2018narrativeqa}, however, we did not observe obvious difference in performance. This is probably because we do not update word embeddings during training.} during training.

\paragraph{Implementation Details} Text is lowercased and tokenized with NLTK\footnote{\url{https://www.nltk.org/}}. For retrieval of paragraphs, we use the cosine similarity between TF-IDF vector representations. TF-IDF representations are vectorized by Scikit-Learn using an N-gram range of $[1,3]$ with stopword filtering. The maximum context size is tuned amongst $\{2000,4000\}$ and reported accordingly. The paragraph/chunk size is dynamic and configured amongst $\{50,100,200,500\}$. The retrieved excerpts are retrieved based on similarity match between context chunks and answer \textbf{or} question depending on the curriculum learning scheme. We tune the maximum answer length amongst $\{6,8,12$\} and the maximum question length is set to $30$. Since two answers are provided for each question, we train on both sets of answers. During construction of the golden labels, first perform an n-gram search of the answer in the context. The largest n-gram match is allocated indices belonging to the context (i.e., [1,$\ell_c$]). For the remainder words, stopwords are automatically allocated indices in the global vocabulary and non-stopwords are assigned context indices. If an answer word is not found, it is ignored. To construct the global vocabulary for the pointer generator decoder and avoid story-specific words, we use words that appear in at least 10 stories. 

\paragraph{Evaluation} During evaluation, we (1) remove the full stop at the end of answers and (2) lowercase both answers. We use the BLEU, Rouge and METEOR scorers provided at \url{https://github.com/tylin/coco-caption}.

\paragraph{Baselines} As baselines, we compare the proposed model with reported results in \cite{kovcisky2018narrativeqa}.. Additionally, we include several baselines which we implement by ourselves. This is in the spirit of providing better (and fairer) comparisons. The compared baselines are listed below:

\begin{itemize}
    \item \textbf{Attention Sum Reader (ASR)} \cite{kadlec2016text} is a simple baseline for reading comprehension. Aside from our the results on \cite{kovcisky2018narrativeqa}, we report our own implementation of the ASR model. Our implementation follows \cite{kovcisky2018narrativeqa} closely.
     \item \textbf{Reinforced Reader Ranker (R$^3$)} \cite{wang2018r} is a state-of-the-art model for open domain question answering, utilizing reinforcement learning to select relevant passages to train the reading comprehension model. Our objective is to get a sense of how well do open-domain models work on understanding narratives.
    \item \textbf{RNET + PG / CPG} \cite{wang2017gated} is a strong, competitive model for paragraph level reading comprehension. We replace the span\footnote{The performance of the RNET + span predictor is similar to the BiDAF model reported in \cite{kovcisky2018narrativeqa}.} prediction layer in RNET with a pointer generator (PG) model with the exact setup as our model. We also investigate equipping RNET + PG with our curriculum learning mechanism (curriculum pointer generator). 

\end{itemize}

\subsection{Experimental Results}
Table \ref{mainresults} reports the results of our approach on the NarrativeQA benchmark. Our approach achieves state-of-the-art results as compared to prior work \cite{kovcisky2018narrativeqa}. When compared to the best ASR model in \cite{kovcisky2018narrativeqa}, the relative improvement across all metrics are generally high, ranging from $+17\%$ to $51\%$. The absolute improvements range from approximately $+1\%$ to $+3\%$.  

Pertaining to the models benchmarked by us, we found that our re-implementation of ASR (Ours) leaves a lot to be desired. Consequently, our proposed IAL-CPG model almost doubles the score on all metrics compared to ASR (Ours). The $R^3$ model, which was proposed primarily for open-domain question answering does better than ASR (Ours) but still fall shorts. Our RNET-PG model performs slightly better than $R^3$ but fails to get a score on BLEU-4. Finally, RNET-CPG matches the state-of-the-art performance of \cite{kovcisky2018narrativeqa}. However, we note that there might be distinct implementation differences\footnote{This is made clear from how our ASR model performs much worse than \cite{kovcisky2018narrativeqa}. We spend a good amount of time trying to reproduce the results of ASR on the original paper. } with the primary retrieval mechanism and environment/preprocessing setup. A good fair comparison to observe the effect of our curricum reading is the improvement between RNET-PG and RNET-CPG.

\subsection{Ablation Study}
In this section, we provide an extensive ablation study on all the major components and features of our proposed model. Table \ref{tab:ablation} reports results of our ablation study. 
\begin{table*}[ht]
  \centering
    \begin{tabular}{c|cccc}
    \hline
    Ablation & B1& B4 & M & RL \\
    \hline
    Original Full Setting & \textbf{23.31} & \textbf{2.70}   & \textbf{5.68}  & \textbf{17.33} \\
    \hline
  
    (1) Remove IAL layer & 18.93  & 1.94  & 4.52  & 14.51 \\
    (2) Replace regular Self-Attention & 19.61  & 0.96 & 4.38  & 15.24 \\
    (3) Remove Enhancement & 20.25  & 1.76 & 4.92  & 15.14 \\
      \hline
    (4) Remove PG + CR & 15.30  & 0.91  & 3.85  & 11.36 \\
    (5) Remove CR (understandability) & 20.13 & 2.30   & 4.94  & 16.96 \\
    (6) Remove CR (answerability) & 20.13 & 1.82  & 4.92  & 15.77 \\
    (7) Train Easy Only & 20.75 & 1.52  & 4.65  & 15.42 \\
    (8) Train Hard Only & 19.18 & 1.49  & 4.60   & 14.19  \\
    \hline
    (9) Add RL & 21.85 & 2.70   & 5.31  & 16.73 \\
    \hline
    (10)  $50 \veryshortarrow 100 \veryshortarrow 200$ &  23.31 & 2.70   & 5.68  & 17.33 \\
    (11) $50 \veryshortarrow 100 \veryshortarrow 200 \veryshortarrow 500$ & 21.07 & 2.86  & 5.33  &16.78 \\
    (12) $100 \veryshortarrow 200 \veryshortarrow 500 \veryshortarrow 50$ & 20.18 & 2.60   & 5.50   & 18.14 \\
    (13) $500 \veryshortarrow 50 \veryshortarrow 100 \veryshortarrow 200$ & 20.95  & 2.51  & 5.41  & 17.05 \\
    (14) $500 \veryshortarrow 200 \veryshortarrow 100 \veryshortarrow 50$ & 17.13 &	2.38 &	4.60 &	15.56 \\
    (15) $50$ (static) & 20.91	& 2.57 & 	5.35	 & 18.78 \\
    (16) $500$ (static) & 19.36 &	2.45 &	4.94 &	16.00 \\
    \hline
    \end{tabular}%
   \caption{Ablation results on NarrativeQA development set. (1-3) are architectural ablations. (4-8) are curriculum reading based ablations. (9) investigates RL-based generation. (10-16) explores the understandability/paragraph size heuristic. Note that (10) was the optimal scheme reported in the original setting. Moreover, more permutations were tested but only representative example are reported due to lack of space.}
   \label{tab:ablation}%
\end{table*}%
\paragraph{Attention ablation} In ablations (1-3), we investigate the effectiveness of the self-attention layer. In (1), we remove the entire IAL layer, piping the context-query layer directly to the subsequent layer. In (2), we replace block-based self-attention with the regular self-attention. Note that the batch size is kept extremely small (e.g., 2), to cope with the memory requirements. In (3), we remove the multiplicative and subtractive features in the IAL layer. Results show that replacing the block-based self-attention with regular self-attention hurts performance the most. However, this may be due to the requirement of reducing the batch size significantly. Removing the IAL layer only sees a considerable drop while removing the enhancement also reduces performance considerably. \paragraph{Curriculum ablation} In ablations (4-8), we investigate various settings pertaining to curriculum learning. In (4), we remove the pointer generator (PG) completely. Consequently,  there is also no curriculum reading in this setting. Performance drops significantly in this setting and demonstrates that the pointer generator is completely essential to good performance. In (5-6), we remove one component from our curriculum reading mechanism. Results show that the answerabiity heuristic is more important than the understandability heuristic. In (7-8), we focus on non curriculum approaches training on the easy or hard set \textbf{only}. It is surprising that training on the hard set alone gives considerablely decent performance which is comparable to the easy set. However, varying them in a curriculum setting has significant benefits. 

\paragraph{RL ablation} In ablation (9), we investigated techniques that pass the BLEU-score back as a reward for the model and train the model jointly using Reinforcement learning. We follow the setting of  \cite{paulus2017deep}, using the mixed training objective and setting $\lambda$ to $0.05$. We investigated using BLEU-1,BLEU-4 and Rouge-L (and combinations of these) as a reward for our model along with varying $\lambda$ rates. Results in Table \ref{tab:ablation} reports the best result we obtained. We found that while RL does not significantly harm the performance of the model, there seem to be no significant benefit in using RL for generating answers, as opposed to other sequence transduction problems \cite{bahdanau2016actor,paulus2017deep}.

\paragraph{Understandability ablation} From ablations (10-16), we study the effect of understandability and alternating paragraph sizes. We find that generally starting from a smaller paragraph and moving upwards performs better and moving the reverse direction may have adverse effects on performance. This is made evident by ablations (10-11). We also note that a curriculum approach beats a static approach often. 

\begin{table*}[htbp]
  \centering
  \small
    \begin{tabular}{p{0.2cm}p{8.0cm}|p{3.3cm}|p{2.8cm}}
    \hline
    \multicolumn{2}{l|}{Question}     & Model Answer & Ground Truth \\
    \hline
    \rowcolor{Gray}
    (1) & how many phases did the court compliment competition have? & two   & 2 \\
    \hline
    (2) & who suffers from a crack addiction? & dick  & dicky \\
    \hline
    \rowcolor{Gray}
    (3) & where did john and sophia go to from the airport? & moscow & russia \\
    \hline
    (4) & what country did nadia's cousin and friend visit her from? & russia & russia \\
    \hline
    \rowcolor{Gray}
    (5) & why is nadia kidnapped by alexei? & to wants be a love of john & because he now wants her to have the baby \\
    \hline
    (6) & who does mary marry? & charles who is her & charles \\
    \hline
    \rowcolor{Gray}
    (7) & what instrument does roberta guaspari play? & violin & violin \\
    \hline
    (8) & where is the school located where roberta takes a position as a substitute violin teacher? & in the york school & east harlem in new york city \\
    \hline
    \rowcolor{Gray}
    (9) & what is the profession of roberta's husband? & she is a naval & he is in the us navy \\
    \hline
    \end{tabular}%
   \caption{Qualitative analysis on NarrativeQA development set.}
   \label{tab:error}%
\end{table*}%

\subsection{Qualitative Error Analysis}

Table \ref{tab:error} provides some examples of the output of our best model. First, we discuss some unfortunate problems with the evaluation in generation based QA. In examples (1), the model predicts a semantically correct answer but gets no credit due to a different form. In (2), no credit is given for word-level evaluation. In (3), the annotators provide a more general answer and therefore, a highly specific answer (e.g., moscow) do not get any credit.

Second, we observe that our model is occasionally able to get the correct (exact match) answer. This is shown in example (4) and (7). However, there are frequent inability to generate phrases that make sense, even though it seems like the model is trudging along the right direction (e.g., ``to wants to be a love of john'' versus ``because he wants her to have the baby'' and ``in the york school'' versus ``east harlem in new york''). In (9), we also note a partially correct anwer, even though it fails to realize that the question is about a male and generates ``she is a naval''. 



\section{Related Work}
The existing work on open domain QA \cite{chen2017reading} has distinct similarities with our problem, largely owing to the overwhelming large corpus that a machine reader has to reason over. In recent years, a multitude of techniques have been developed. \cite{wang2018r} proposed reinforcement learning to select passages using the reader as the reward. \cite{min2018efficient} proposed ranking the minimal context required to answer the question. \cite{clark2017simple} proposed shared norm method for predicting spans in the multi-paragraph reading comprehension setting. \cite{lin2018denoising} proposed ranking and de-noising techniques. \cite{wang2017evidence} proposed evidence aggregation based answer re-ranking. Most techniques focused on constructing a conducive and less noisy context for the neural reader. Our work provides the first evidence of diverse sampling for training neural reading comprehension models. 

Our work draws inspiration from curriculum learning (CL) \cite{bengio2009curriculum}. One key difficulty in CL is to determine which samples are easy or hard. Self-paced learning \cite{jiang2015self} is a recently popular form of curriculum learning that treats this issue as an optimization problem. To this end, \cite{sachan2016easy} applies self-paced learning for neural question answering. Automatic curriculum learning \cite{graves2017automated}, similarly, extracts signals from the learning process to infer progress.

State-of-the-art neural question answering models are mainly based on cross-sentence attention \cite{seo2016bidirectional,wang2016machine,xiong2016dynamic,tay2018densely}. Self-attention \cite{vaswani2017attention,wang2017gated} has also been popular for reading comprehension \cite{wang2018r,clark2017simple}. However, its memory complexity makes it a challenge for reading long context. Notably, the truncated/summary setting of the NarrativeQA benchmark have been attempted recently \cite{tay2018densely,tay2018recurrently,hu2018attention,tay-etal-2018-multi}. However, this summary setting bypasses the difficulties of long context reading comprehension, reverting to the more familiar RC setup. 

While most of the prior work in this area has mainly focused on span prediction models \cite{wang2016machine} and/or multiple choice QA models \cite{wang2016compare}, there have been recent interest in generation based QA \cite{tan2017s}. S-NET \cite{tan2017s} proposed a two-stage retrieve then generate framework. 

Flexible neural mechanisms that learn to point and/or generate have been also popular across many NLP tasks. Our model incorporates Pointer-Generator networks \cite{see2017get} which learns to copy or generate new words within the context of neural summarization. Prior to Pointer Generators, CopyNet \cite{gu2016incorporating} incorporates a copy mechanism for sequence to sequence learning. Pointer generators have also been recently adopted for learning a universal multi-task architecture for NLP \cite{mccann2018natural}.

\section{Conclusion}
We proposed curriculum learning based Pointer-generator networks for reading long narratives. Our proposed IAL-CPG model achieves state-of-the-art performance on the challenging NarrativeQA benchmark. We show that sub-sampling diverse views of a story and training them with a curriculum scheme is potentially more effective than techniques designed for open-domain question answering. We conduct extensive ablation studies and qualitative analysis, shedding light on the task at hand.

\bibliography{acl2019}
\bibliographystyle{acl_natbib}

\end{document}